\documentclass[sigconf]{acmart}
\usepackage{multirow}
\copyrightyear{2022}
\acmYear{2022}
\setcopyright{acmcopyright}
\acmConference[MM '22] {Proceedings of the 30th ACM International Conference on Multimedia }{October 10--14, 2022}{Lisbon, Portugal.}
\acmBooktitle{Proceedings of the 30th ACM International Conference on Multimedia (MM '22), October 10--14, 2022, Lisbon, Portugal}
\acmPrice{15.00}
\begin{document}
\fancyhead{}

\title{Towards Counterfactual Image Manipulation via CLIP}

\author{Yingchen Yu}
\email{yingchen001@e.ntu.edu.sg}
\affiliation{%
  \institution{Nanyang Technological University \& Alibaba Group}
  \country{Singapore}
}
\author{Fangneng Zhan}
\email{fzhan@mpi-inf.mpg.de}
\affiliation{%
  \institution{Max Planck Institute for Informatics}
  \city{Saarbr\"ucken}
  \state{Saarland}
  \country{Germany}
}
\author{Rongliang Wu}
\email{ronglian001@e.ntu.edu.sg}
\affiliation{%
  \institution{Nanyang Technological University}
  \country{Singapore}
}
\author{Jiahui Zhang}
\email{ronglian001@e.ntu.edu.sg}
\affiliation{%
  \institution{Nanyang Technological University}
  \country{Singapore}
}
\author{Shijian Lu}
\authornote{Corresponding author.}
\email{shijian.lu@ntu.edu.sg}
\affiliation{%
  \institution{Nanyang Technological University}
  \country{Singapore}
}
\author{Miaomiao Cui}
\email{miaomiao.cmm@alibaba-inc.com}
\affiliation{%
  \institution{DAMO Academy, Alibaba Group}
  \city{Beijing}
  \state{}
  \country{China}
}
\author{Xuansong Xie}
\email{xingtong.xxs@taobao.com}
\affiliation{%
  \institution{DAMO Academy, Alibaba Group}
  \city{Beijing}
  \state{}
  \country{China}
}
\author{Xian-Sheng Hua}
\email{xshua@outlook.com}
\affiliation{%
  \institution{Zhejiang University}
  \city{Hangzhou}
  \state{Zhejiang}
  \country{China}
}
\author{Chunyan Miao}
\email{ascymiao@ntu.edu.sg}
\affiliation{%
  \institution{Nanyang Technological University}
  \country{Singapore}
}
\renewcommand{\shortauthors}{Yu, et al.}
\begin{abstract}
Leveraging StyleGAN's expressivity and its disentangled latent codes, existing methods can achieve realistic editing of different visual attributes such as age and gender of facial images. An intriguing yet challenging problem arises: Can generative models achieve counterfactual editing against their learnt priors? Due to the lack of counterfactual samples in natural datasets, we investigate this problem in a text-driven manner with Contrastive-Language-Image-Pretraining (CLIP), which can offer rich semantic knowledge even for various counterfactual concepts. Different from in-domain manipulation, counterfactual manipulation requires more comprehensive exploitation of semantic knowledge encapsulated in CLIP as well as more delicate handling of editing directions for avoiding being stuck in local minimum or undesired editing. To this end, we design a novel contrastive loss that exploits predefined CLIP-space directions to guide the editing toward desired directions from different perspectives. In addition, we design a simple yet effective scheme that explicitly maps CLIP embeddings (of target text) to the latent space and fuses them with latent codes for effective latent code optimization and accurate editing. Extensive experiments show that our design achieves accurate and realistic editing while driving by target texts with various counterfactual concepts.
\end{abstract}



\begin{teaserfigure}
  \includegraphics[width=0.98\textwidth]{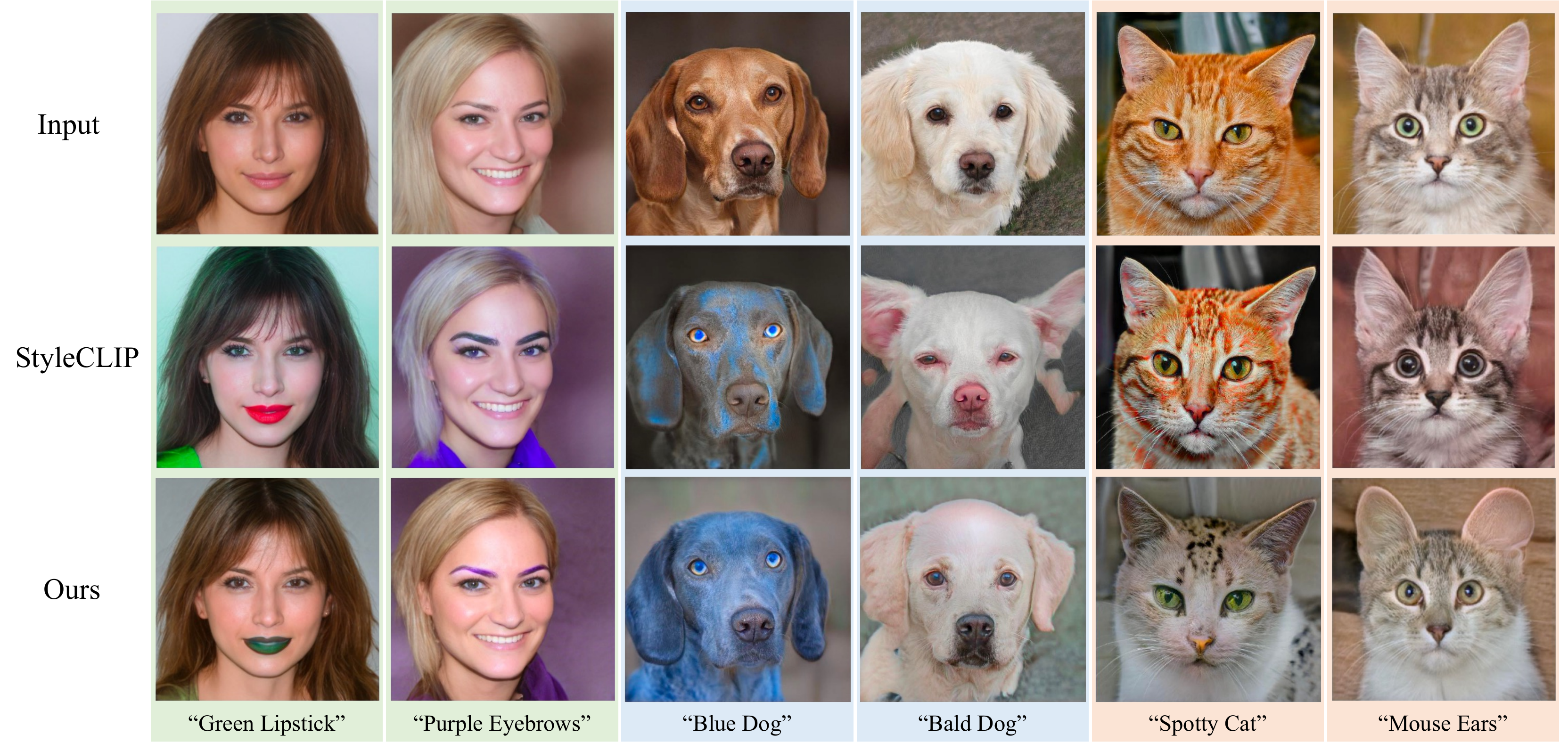}
  \caption{
  Illustration of text-driven counterfactual manipulation: The state-of-the-art StyleCLIP \cite{patashnik2021styleclip} often struggles to meet target counterfactual descriptions as it 
  optimizes CLIP scores directly, which is susceptible to adversarial solutions.
  Our design achieves more accurate and robust counterfactual editing via effective text embedding mapping and a novel contrastive loss CLIP-NCE that comprehensively exploits semantic knowledge of CLIP.
  }
  \label{fig:teaser}
\end{teaserfigure}

\maketitle

\section{Introduction}

Empowered by Generative Adversarial Networks (GANs)~\cite{goodfellow2014generative}, we have observed great advances in image manipulation in recent years~\cite{zhu2016generative, wu2020cascade, wu2020leed, zhan2021bi, xia2021tedigan, patashnik2021styleclip}. In particular, StyleGANs~\cite{karras2019style, karras2020analyzing, karras2021alias} keep extending the boundary of realistic image synthesis. Meanwhile, their learnt latent spaces possess disentanglement properties, enabling various image manipulations by directly editing the latent codes of pretrained StyleGAN~\cite{harkonen2020ganspace, shen2020interfacegan, abdal2021styleflow, wu2021stylespace}.

As StyleGAN is pretrained with images of a specific domain, most related manipulation methods are restricted to certain domain-specific editing, e.g., changing hair styles for facial dataset. The conventional way of moving beyond the domain is to retrain the model with task-specific samples, but it becomes infeasible in counterfactual concept generation where training sample are scarce. Eliminating the need of manual efforts or additional training data has recently been explored for StyleGAN-based image manipulation. For example, StyleCLIP \cite{patashnik2021styleclip} achieves text-guided image manipulation by editing StyleGAN latent codes under the sole supervision of CLIP. However, StyleCLIP optimizes CLIP scores directly, which is susceptible to adversarial solutions or get stuck in local minimum \cite{liu2021fusedream}. Thus, it's typically restricted to in-domain manipulation. StyleGAN-NADA~\cite{gal2021stylegan} enables out-of-domain generation with directional CLIP loss, which encourages generation diversity and avoids adversarial solutions by aligning CLIP-space directions between source and target text-image pairs. However, StyleGAN-NADA optimizes the entire generator for zero-shot domain adaptation, which often requires manual setting of optimization iterations for different cases to maintain manipulation accuracy and quality. 
Additionally, the directional CLIP loss guides the latent code optimization with only the identical CLIP-space directions (source to target text embedding), which does not explicitly regularize the editing strength. Thus, it is prone to allow the model to over-edit the latent codes, and consequently lead to inaccurate or excessive editing.

We design \textbf{CF-CLIP}, a \textbf{CLIP}-based text-guided image manipulation network that allows accurate \textbf{C}ounter\textbf{F}actual editing without requiring additional training data or optimization of the entire generator. For counterfactual editing against the learnt prior, it is indispensable to extract semantic knowledge of CLIP for discovering the editing direction in the latent space of a pretrained GAN. To this end, we design a CLIP-based Noise Contrastive Estimation (CLIP-NCE) loss that explores the CLIP semantic knowledge comprehensively. Instead of only emphasizing the identical directions as in directional CLIP loss \cite{gal2021stylegan}, CLIP-NCE maximizes the mutual information of selected positive pairs while minimizing it along other directions (i.e. negative pairs) in the CLIP space, which yields auxiliary information and facilitates latent code optimization toward desired editing directions. 

In addition, editing accuracy is essential for manipulation tasks, which means that manipulations should only take place in the target edit regions and faithfully reflect the target description. In this paper, we enhance the editing accuracy in two different ways. First, we augment the target images with different perspective views before computing CLIP-NCE. This introduces two benefits: 1) the augmentations preserve semantic information and prevent the model from converging to adversarial solutions; 2) the consistent semantics across different perspective views encourage the model to discover the underneath geometry information in CLIP \cite{kwon2021clipstyler}, which helps for better understanding of image semantics as well as precise editing locally or globally.
Second, we design a simple yet effective text embedding mapping (TEM) module to explicitly utilize the semantic knowledge of CLIP embeddings. The CLIP-space embedding of target text is separately mapped into the StyleGAN latent space to disentangle the semantics of text embeddings. After fusing the disentangled text embeddings with input latent codes, the model could leverage the semantic information to highlight the target-related latent codes in relevant StyleGAN layers. 
The TEM thus enables the model to accurately locate the target editing regions and effectively suppress unwanted editing.

The contributions of this work can be summarized in three aspects. 
First, we design CF-CLIP, a CLIP-based image manipulation framework that enables accurate and high-fidelity counterfactual editing given target textual description.
Second, we design a contrastive loss in the CLIP space, dubbed CLIP-NCE, which provides comprehensive guidance and enables faithful counterfactual editing. 
Third, we design a simple yet effective text embedding mapping module (TEM) that allows explicit exploitation of CLIP embeddings during latent code optimization to facilitate accurate editing.

\section{Related Work}
\subsection{Text-guided synthesis and manipulation}
Most existing work in text-guided synthesis adopts conditional GAN \cite{mirza2014conditional} that treats text embedding as conditions \cite{zhan2021multimodal}. For example, \citet{reed2016generative} extracts text embeddings from a pretrained encoder. \citet{xu2018attngan} employs an attention mechanism to improve the generation quality. Instead of training GANs from scratch, TediGAN \cite{xia2021tedigan} incorporates pretrained StyleGAN for text-guided image synthesis and manipulation by mapping the text to the latent space of StyleGAN.  In addition, a large-scale pretrained CLIP model \cite{radford2021learning} for joint vision-language representation has recently been released. 
With powerful CLIP representations, StyleCLIP \cite{patashnik2021styleclip} achieves flexible and high-fidelity manipulation by exploring the latent space of StyleGAN. Instead of confining generated images to the trained domain, CLIPstyler~\cite{kwon2021clipstyler} eliminates the necessity of StyleGAN and allows for text-guided style transfer with arbitrary source images. Beyond that, StyeGAN-NADA~\cite{gal2021stylegan} optimizes the pretrained generative model with text conditions only, which enables out-of-domain manipulation via domain adaptation. As optimizing latent codes for counterfactual manipulation is more challenging compared to in-domain manipulation, the optimization of latent codes tend to get stuck in local minimal or excessive modification during training and lead to inaccurate editing or artifacts. 
Our model can mitigate this issue effectively by providing comprehensive guidance in the CLIP-space with the proposed CLIP-NCE loss.

\subsection{Counterfactual image generation}
The early study of counterfactual image generation serves as a tool for explaining image classifier. For example, \citet{chang2018explaining} presents generative infilling to produce counterfactual contents within masked regions for aligning with the data distributions.
For robust out-of-domain classification, \citet{sauer2021counterfactual} enables counterfactual generation by disentangling object shape, texture, and background without direct supervision. However, these counterfactual generations often lack fidelity and consistency around the mask boundary. This situation is greatly alleviated with the appearance of Dall-E \cite{ramesh2021zero}, which has billions of parameters and is trained with large-scale text-image pairs. With such powerful expressivity, Dall-E enables high-fidelity generation with a prompting text, even with novel counterfactual concepts. Leveraging the power of the large-scale pretrained model CLIP \cite{radford2021learning}, \citet{liu2021fusedream} present a training-free, zero-shot framework that can generate counterfactual images. Similarly, \citet{gal2021stylegan} introduce a zero-shot domain adaptation method that enables out-of-domain generation even when the target domain is counterfactual. However, the aforementioned methods mainly focus on the global style or semantic changes when generating counterfactual images, while our proposed method could achieve more flexible and coherent counterfactual manipulation both locally and globally.

\section{Preliminaries}
Similar to most existing CLIP-based manipulation methods \cite{patashnik2021styleclip,gal2021stylegan}, our method also leverages StyleGAN and CLIP, where the former yields disentangled latent codes and high-fidelity generation, and the latter provides powerful semantic knowledge for guiding the editing direction. We therefore first briefly review the fundamentals of StyleGAN and CLIP and highlight their relevant properties that allow counterfactual image editing.
\subsection{StyleGAN}
StyleGANs \cite{karras2019style,karras2020analyzing} are unconditional generative models that can synthesize high resolution images progressively from random noises. Specifically, random noise $z \in \mathbb{R}^{512}$ is sampled from a Gaussian distribution and injected into a mapping network, which projects $z$ into a learned latent space $W \in \mathbb{R}^{512}$. Consequently, the latent codes $w$ are injected into different layers of the synthesis network to control different semantics at different resolutions, which endows $W$ space with good disentanglement properties during training \cite{collins2020editing,shen2020interfacegan}. Recent GAN-inversion studies present $\mathcal{W+} \in \mathbb{R}^{{n\_latent} \times 512}$ space for fine-grained control of image semantics \cite{abdal2019image2stylegan,richardson2021encoding,tov2021designing,alaluf2021restyle}, where $n\_latent$ is the number of StyleGAN layers. Although the synthesis generator is typically constrained by the strong prior learned from the training domain, the disentanglement of the latent codes suggests the feasibility of recombining different attributes to achieve counterfactual or out-of-domain manipulation. 
However, discovering the proper editing directions remains a challenge. One promising solution is to incorporate the guidance from large-scale CLIP based on its powerful text-image representation.

\subsection{CLIP}
\label{section:clip}
CLIP \cite{radford2021learning} is trained with over 400 million text-image pairs to learn a joint representation of vision-language through contrastive learning. It consists of two encoders that encode images and texts into 512-dimensional embedding, respectively. By minimizing the cosine distance between encoded embeddings of large-scale text-image pairs, CLIP learns a powerful multi-modal embedding space where the semantic similarity between texts and images can be well measured. Thanks to large-scale training, rich semantic knowledge is encapsulated in CLIP embedding, which could guide counterfactual or out-of-domain generation for pretrained generators. However, effective exploitation of such powerful representations remains an open research problem, which is also the focus of our study.

\subsubsection{Global CLIP Loss}

Recently, StyleCLIP \cite{patashnik2021styleclip} directly utilizes the CLIP-space cosine distance to guide image manipulation according to the semantic of target text, which can be formulated by a global CLIP loss:
\begin{equation}
    \mathcal{L}_{global} = 1-cos(E_I(I_{edit}), E_T(t_{tgt})) ,
\end{equation}
where $cos$ stands for cosine similarity, $E_I$ and $E_T$ denote the image encoder and text encoder of CLIP, respectively, $I_{edit}$ is the manipulated image, and $t_{tgt}$ is the target textual description. 

\subsubsection{Directional CLIP Loss}
A known issue of global CLIP loss is adversarial solutions \cite{liu2021fusedream}, which means that the model tends to fool the CLIP classifier by adding meaningless pixel-level perturbations to the image \cite{gal2021stylegan}. To mitigate such issue, a directional CLIP loss \cite{gal2021stylegan} is proposed to align the CLIP-space directions between the source and target text-image pairs.
, which is defined as:
\begin{align}
& \Delta  T = E_T(t_{tgt}) - E_T(t_{src}), \nonumber \\
& \Delta  I = E_I(I_{edit}) - E_I(I_{src}), \nonumber \\
& \mathcal{L}_{dir} = 1 - cos(\Delta  T, \Delta  I), 
\end{align} 
where $I_{src}=G(w)$ is the source image, and $t_{src}$ denotes the source texts with neutral descriptions such as "Photo", "Face", and "Dog".

\begin{figure*}[t]
    \centering
    \includegraphics[width=0.89\linewidth]{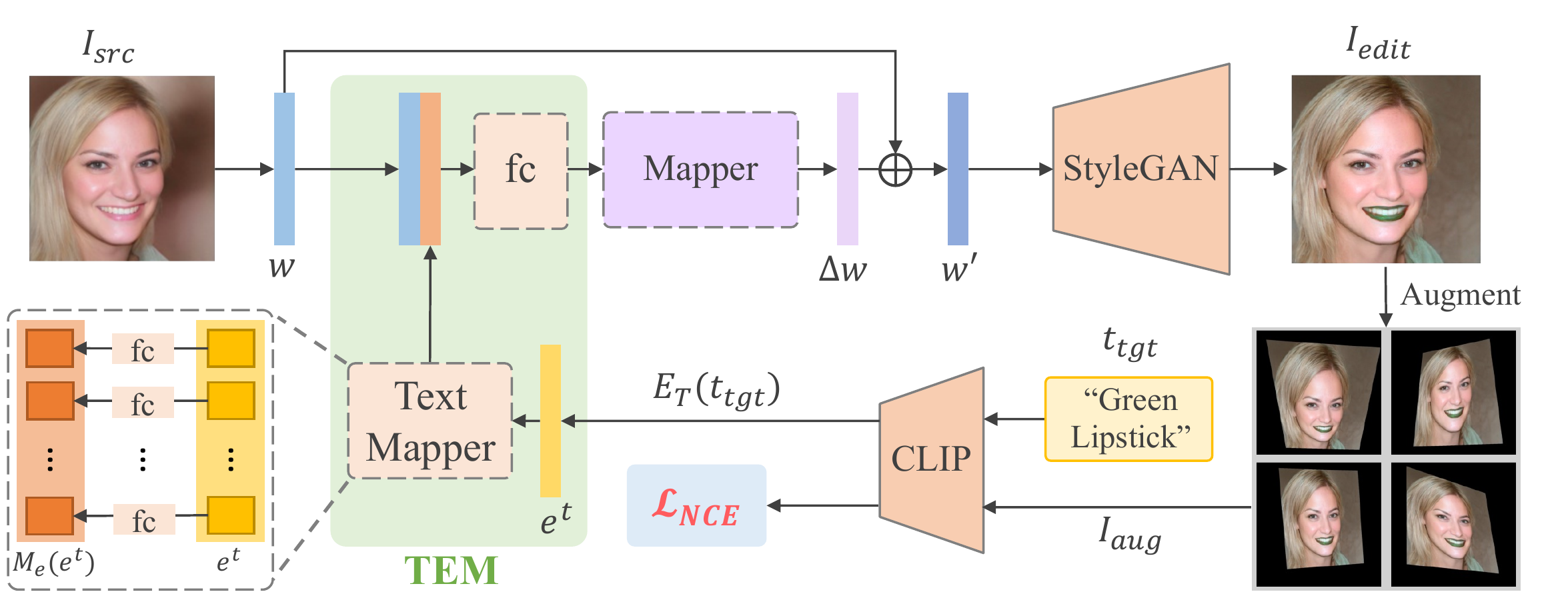}
    \caption{
    The framework of the proposed text-driven counterfactual image manipulation: Given a randomly sampled or inverted latent code $w$, a \textit{Text Mapper} (detailed structures shown at the bottom left) first projects the CLIP embedding of the target text $t_{tgt}$ (i.e., $e^t$) to a latent space and fuses it with $w$ to highlight the editing in relevant generator layers. The fused embedding is then propagated to a \textit{Mapper} to yield residual $\Delta w$ for editing. With the manipulated latent code $w'$, manipulated images are generated by StyleGAN which are further augmented and encoded in the CLIP space for loss calculation. We exploit the semantic knowledge in CLIP to discover the editing direction under the guidance of the proposed \textit{CLIP-NCE} loss $\mathcal{L}_{NCE}$.
    } 
    \label{fig_arch}
\end{figure*}

\section{Method}
Given a text prompt that describes the target counterfactual manipulation together with a source image, our goal is to faithfully produce manipulated images against the strong prior of the pretrained generator. Due to the lack of counterfactual examples in natural dataset, we achieve this goal in a zero-shot manner by leveraging the semantic knowledge of the pretrained model CLIP as the sole supervision. Since manipulating the latent codes for counterfactual manipulation is typically more challenging than in-domain manipulation, how to comprehensively extract the semantic information encapsulated in CLIP and carefully guide the editing directions in the latent space are focal points of our approach. 

\subsection{Pipeline}
The framework of our proposed approach is illustrated in Fig. \ref{fig_arch}. With the target text prompt $t_{tgt}$, the text embeddings $E_T(t_{tgt})$ extracted by the text encoder of CLIP is firstly forwarded to the Text Embedding Mapping (TEM) module, which is trained to project the encoded text embeddings into the latent space for more explicit guidance from the CLIP model. Subsequently, the projected feature is concatenated with the latent code $w$, which can be randomly generated or inverted from real images. The concatenated latent codes is then fed into a fully-connected layer for feature fusion. Similarly to StyleCLIP \cite{patashnik2021styleclip}, we employ a mapper to obtain the residual $\Delta w$ from the fused features. After summing it back to the source latent code $w$, the resulting $w'=w+\Delta w$ is passed to the synthesis network of StyleGAN to produce the manipulated images $I_{src}$. Next, the manipulated images $I_{src}$ are augmented with different perspective views, and the resulting images $I_{aug}$ are used to calculate the proposed CLIP-NCE loss.

\subsection{CLIP-NCE}
Inspired by CLIPStyler \cite{kwon2021clipstyler}, a random perspective augmentation is applied to the manipulated image $I_{edit}$ before calculating the CLIP-NCE loss, and this augmentation scheme facilitates our framework in approaching its goal from two perspectives. First, it becomes harder for the model to fool the CLIP with adversarial solutions, because now it has to simultaneously produce appropriate perturbations across most of the randomly augmented images. Second, we conjecture that during the large-scale pretraining process, CLIP may learn to model the geometry information of the same object under different views. Hence, the multiple views provided by the perspective augmentation yield CLIP representations with geometry information for different views, which help the model explore semantic information of the CLIP model in a 3D-structure-aware manner.

Therefore, the augmented images can be formulated as:
\begin{equation}
    I_{aug} = aug(I_{edit}),
\end{equation}
where $aug(\cdot)$ is the random perspective augmentation.

As mentioned in Section \ref{section:clip}, StyleCLIP \cite{patashnik2021styleclip} adopted global CLIP loss for text-guided manipulation. Despite the adversarial solutions issue, global CLIP loss may also lead to a local minimal issue during optimization. Taking "Green Lipstick" as an example, the global CLIP loss could be stuck at a local minimum such that the model changes other more entangled areas into green (e.g., hair), other than the target "lipstick". The directional CLIP loss \cite{gal2021stylegan} helps to mitigate this issue by aligning the CLIP-space directions between the text-image pairs of source and target. However, the CLIP-space directions from source to target texts are almost fixed, directional CLIP loss may over-emphasize the identical directions and neglect the other useful information encapsulated in the CLIP-space embeddings, which tends to allow the latent codes to travel too far and result in excessive or inaccurate editing.

\begin{figure}[t]
    \centering
    \includegraphics[width=0.9\linewidth]{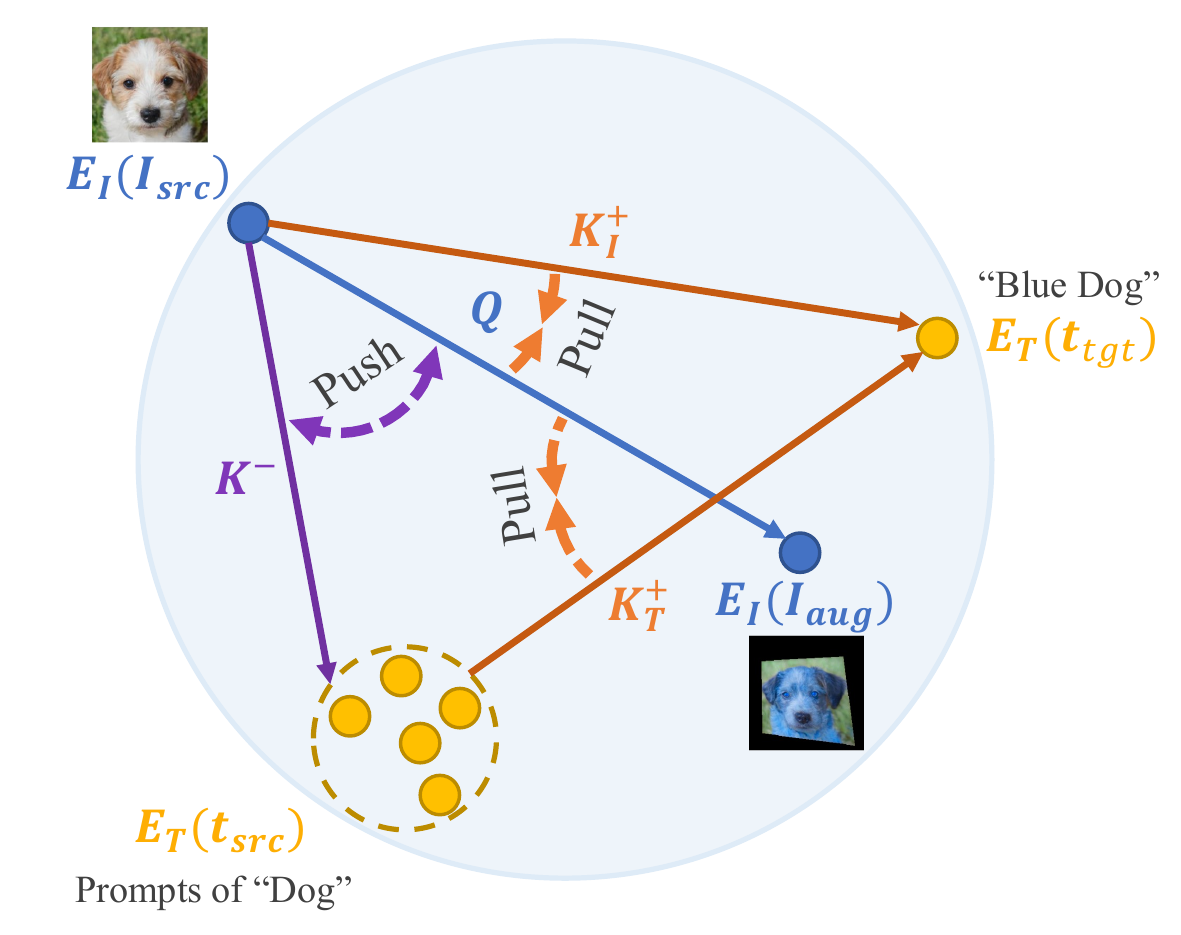}
    \caption{
    Illustration of \textit{CLIP-NCE}: The contrastive loss is computed in the CLIP space, where the direction from the source to the manipulated embedding (blue arrow) forms the \textit{query} for optimization. Two types of \textit{positive samples} are used (orange arrows), where $K_T^+$ encourages the query to align with the direction from source to target texts and $K_I^+$ regularizes the latent codes from traveling too far. The purple arrow highlights \textit{negative samples} from the source image to prompts of source text. By pulling positive pairs and pushing negative pairs, CLIP-NCE enables comprehensive exploitation of CLIP representations.}
    
    \label{fig_nce}
\end{figure}

Therefore, we propose a novel CLIP-based Noise Contrastive Estimation (CLIP-NCE) to maximize/minimize the mutual information between positive/negative pairs, which allows us to leverage the semantic information of the CLIP model in a more comprehensive manner. In order to apply the CLIP-NCE loss, we need to define the query, positive and negative samples like common contrastive loss \cite{van2018representation, zhang2021blind, zhan2022modulated, zhan2022marginal}. The CLIP-NCE then pulls positive samples towards the query $Q$ and pushes negative samples far away from it, as illustrated in Fig. \ref{fig_nce}. 
Firstly, we define the query of the contrastive loss as follows:
\begin{equation}
    Q = E_I(I_{aug}) - E_I(I_{src}),
\end{equation}
$Q$ represents the CLIP-space direction from augmented image to source image, and it is the only term that reflects the optimization of the network during training. Next, positive samples are formulated as follows, which consist of two components:
\begin{align}
K^+_T & = E_T(t_{tgt}) - E_T(t_{src}), \nonumber\\
K^+_I & = E_T(t_{tgt}) - E_I(I_{src}).
\end{align}
The first term $K^+_T$ represents the CLIP-space direction from target text to source text. Similarly to directional CLIP loss \cite{gal2021stylegan}, it is used to regularize the editing direction to align with the text embedding direction from source text to target text. The second term $K^+_I $ is defined as the CLIP-space direction from source image to target text. Sharing the same minus term $- E_I(I_{src}) $ with $Q$, the positive samples encourage the editing directions pointing to the target text embedding, which helps to regularize the editing strength of the latent codes when the model over-emphasizes the text embedding direction $K^+_T$. 

To best distill the semantic information of CLIP model, leveraging as many as useful information in CLIP-space may help to provide more comprehensive guidance. Hence, we define negative samples $K^-$ as the CLIP-space directions from source image to various neural text descriptions, which may prevent the model from producing lazy manipulations that are classified as neutral descriptions like the source text. 

Hence, we design the negative samples $K^-$ to avoid the editing toward the CLIP embeddings with neutral text descriptions:
\begin{equation}
    K^- = E_T(t_{src}) - E_I(I_{src})
\end{equation}
To increase the diversity of source texts, we adopt prompt engineering \cite{radford2021learning}  to fill $t_{src}$ in text templates such as "a photo of a \{Dog\}."

According to \cite{park2020contrastive,andonian2021contrastive}, maximizing mutual information allows diverse yet plausible solutions and helps to avoid averaged solutions for different instances. 

Therefore, we adopt InfoNCE loss \cite{van2018representation} for the carefully defined terms. By maximizing the mutual information between the two selected positive pairs and minimizing the mutual information between negative pairs in the CLIP-space, CLIP-NCE loss provides a comprehensive guide of CLIP and significantly facilitates the optimization of latent codes. Specifically, the proposed CLIP-NCE loss can be formulated as follows:
\begin{eqnarray}
\label{infonce}
\mathcal{L}_{NCE} =& - \log
\frac{e^{(Q \cdot K^+_T / \tau)}} 
{e^{(Q \cdot K^+_T / \tau)} + \sum_{K^-} e^{(Q \cdot K^- / \tau)}}  \nonumber\\
& - \log
\frac{e^{(Q \cdot K^+_I / \tau)}} 
{e^{(Q \cdot K^+_I / \tau)} + \sum_{K^-} e^{(Q \cdot K^- / \tau)}} ,
\end{eqnarray}
where $\tau$ is the temperature and and is set to 0.1 in our approach.  
\begin{figure*}[t]
    \centering
    \includegraphics[width=0.955\linewidth]{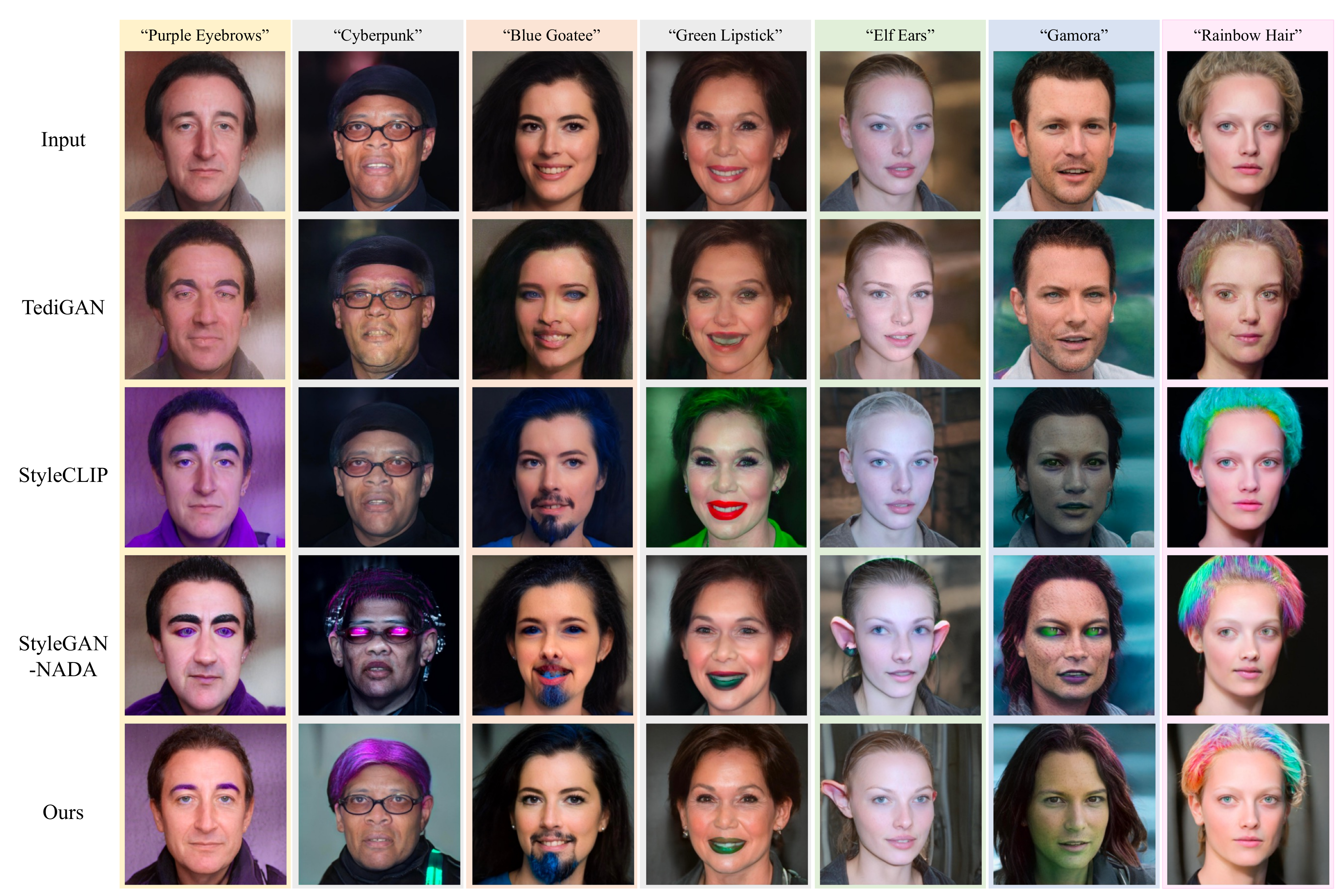}
    \caption{
    Qualitative comparisons with the state-of-the-art over CelebA-HQ \cite{karras2017progressive}: Rows 1-2 show target texts for counterfactual manipulation and input images respectively, and the rest rows show manipulations by different methods. The proposed CF-CLIP achieves clearly better visual photorealism and more faithful manipulation with respect to the target texts.
    }
    \label{comp_face}
\end{figure*} 
\subsection{Text Embedding Mapping}
To enhance editing accuracy by explicitly incorporating the semantic knowledge of CLIP embeddings, we design a simple yet effective text embedding mapping (TEM) module 
to map the CLIP-space embeddings of target text $t_{tgt}$ into the latent space and fuse them with the original latent code $w$. Given
$t_{tgt}$, we first embed it into a 512-dim vector in CLIP space with the text encoder $E_T$ of CLIP, i.e. $e^t=E_T(t_{tgt})$. 
To disentangle text embeddings and allows key editing directions to be propagated to the corresponding generator layers, we employ a text mapper consisting of $n\_latent$ mapping networks, each of which has 4 consecutive fully-connected layers and projects text embeddings into the latent space corresponding to its layer. The text mapper can be defined as:
\begin{equation}
M_e(e^t) = (M^1_e(e^t), M^2_e(e^t),\dots ,M^{n\_latent}_e(e^t)),
\end{equation}
where $n\_latent$ refers to the number of StyleGAN layers. Afterward, we concatenate the projected embedding with the original latent code $w$ and fuse it with a fully-connected layer $M_f$ to obtain the fused embedding that lies in the latent space of StyleGAN:
\begin{equation}
e^f = M_f(M_e(e^t)\oplus w),
\end{equation}
where $\oplus$ is the concatenate operation. Lastly, we apply the mapper of StyleCLIP \cite{patashnik2021styleclip} to obtain the modified latent code:
\begin{equation}
w' = w + M_t(e^f),
\end{equation}
where $M_t$ is the mapper that consists of 3 groups (coarse, medium, and fine) of mapping networks to yield the target latent codes.

\subsection{Loss Functions}
In addition to the proposed CLIP-NCE loss, we follow SyleCLIP \cite{patashnik2021styleclip}, to employ a latent norm loss and a identity loss for face-related manipulation to regularize the editing so that the identity could be persevered. The latent norm loss is the $L_2$ distance in latent space, i.e. $\mathcal{L}_{L2}=||w-w'||_2$. Identity loss uses the pretrained face recognition model, ArcFace $R$ \cite{deng2019arcface}, which is defined as:
\begin{equation}
\mathcal{L}_{ID} = 1 - cos(R(I_{edit}), R(I_{src}))
\end{equation}
As the identity loss does not work for non-facial datasets, We employ a perceptual loss \cite{johnson2016perceptual} to penalize the semantic discrepancy:
\begin{equation}
    \mathcal{L}_{perc} = ||\Phi(I_{edit}) - \Phi(I_{src})||_{1},
\end{equation}
where $\Phi$ is the activation of the \textit{relu5\_2} layer of the VGG-19 model.

Therefore, the overall loss function of our approach is formulated as a weighted combination of the aforementioned losses.
\begin{equation}
    \mathcal{L} = \lambda_{NCE}\mathcal{L}_{NCE} + \lambda_{L2}\mathcal{L}_{L2} + \lambda_{ID}\mathcal{L}_{ID} + \lambda_{perc}\mathcal{L}_{perc},
\end{equation}
where $\lambda_{NCE}$, $\lambda_{L2}$, $\lambda_{ID}$ and $\lambda_{perc}$ are empirically set at 0.3, 0.8, 0.2 and 0.01, respectively, in our implementation. Note that we set $\lambda_{ID}$ to 0 for non-facial datasets and $\lambda_{perc}$ to 0 for facial datasets.

\section{Experiments}

\subsection{Settings}
\subsubsection{Datasets} $\\$
\noindent
$\bullet$ FFHQ \cite{karras2019style}: It consists of 70,000 high-quality face images crawled from Flickr with considerable variations. We adopt a StyleGAN model pretrained over this dataset as our synthesis generator for facial image manipulation. 

\noindent
$\bullet$ CelebA-HQ \cite{karras2017progressive}: It is a high-quality version of the human face dataset CelebA \cite{liu2015deep} with 30,000 aligned face images. We follows the StyleCLIP \cite{patashnik2021styleclip} to train and test our model on the inverted latent codes of CelebA-HQ face images with e4e \cite{tov2021designing} model.

\noindent
$\bullet$ AFHQ \cite{choi2020stargan}: It contains high quality aligned animal images. The dataset provides 3 domains of cat, dog, and wild life, each yielding 5,000 images. We conduct experiments on cat and dog domains with the corresponding pretrained StyleGAN models.

\subsubsection{Compared Methods} $\\$
\noindent
$\bullet$ TediGAN \cite{xia2021tedigan} is the state-of-the-art text-driven face generation and manipulation method. It performs latent optimization for each latent code and target text. We set the number of optimization iterations at 200 based on official TediGAN implementation, and conduct comparisons on face-related manipulation only.

\noindent
$\bullet$ StyleCLIP \cite{patashnik2021styleclip} is the first text-drive image manipulation method that leverages the power of CLIP. It presents 3 different techniques for image manipulation. We applied the StyleCLIP mapper to conduct comparisons with the official implementation.

\noindent
$\bullet$ StyleGAN-NADA \cite{gal2021stylegan} enables text-driven, out-of-domain generator adaptation with CLIP. As it is almost impossible to find the optimal setting for each case, we empirically set the number of optimization iterations at 300 as the image quality becomes worse afterwards. A StyleCLIP mapper is trained for each case based on the shifted generator following the official implementation. 
\begin{figure}[t]
    \centering
    \includegraphics[width=1.0\linewidth]{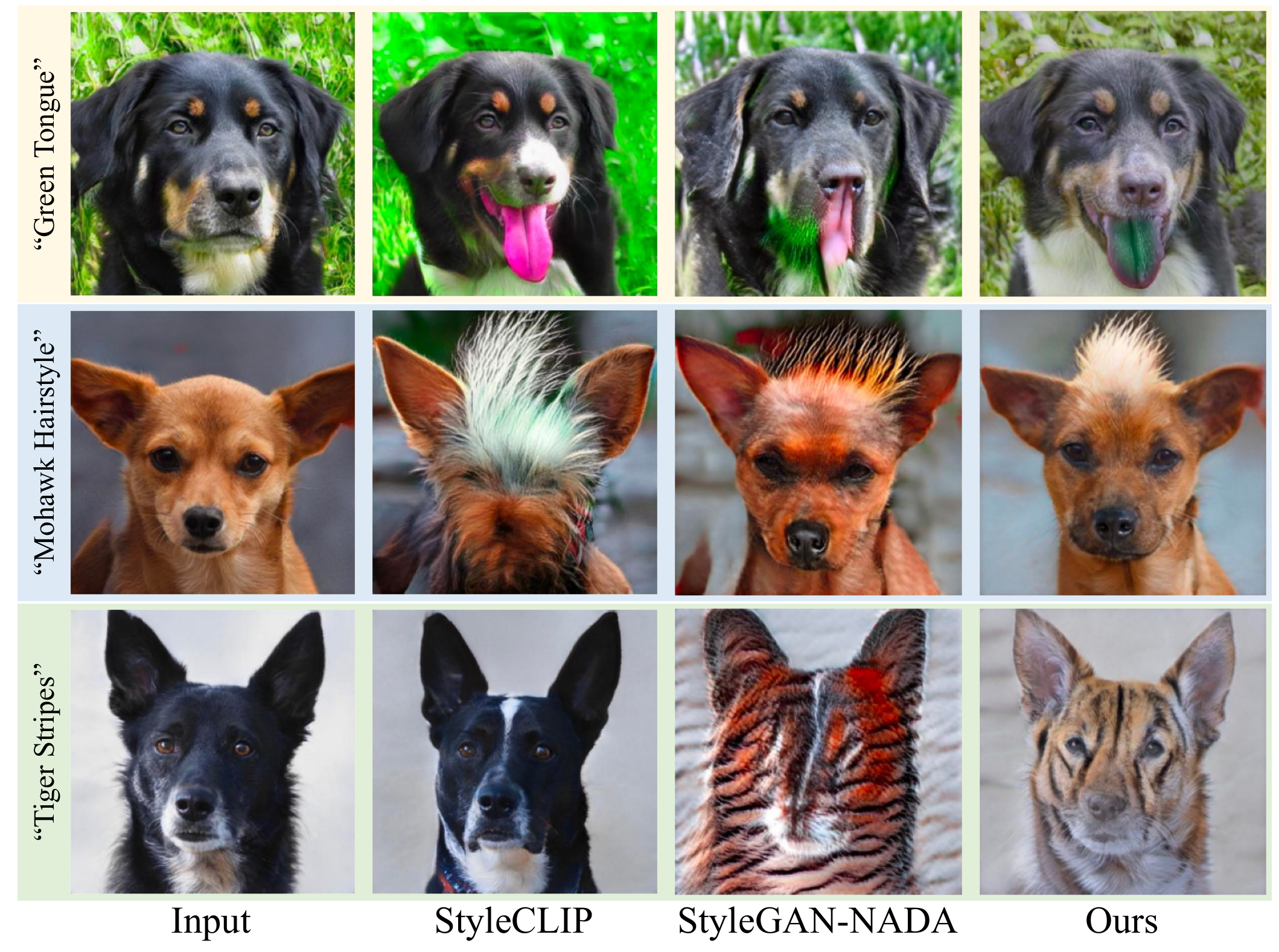}
    \caption{Visual comparison with the state-of-the-art over AFHQ Dog \cite{choi2020stargan}. The target texts appear on the left column and the input images are sampled randomly.}
    \label{comp_dog}
\end{figure}

\subsubsection{Implementation Details}
We trained on inverted latent codes of CelebA-HQ \cite{karras2017progressive} for facial-related manipulation and randomly sampled latent codes for other domains. 
We adopt Adam \cite{kingma2014adam} optimizer with a learning rate of 0.5 as in \cite{patashnik2021styleclip}. The number of training iterations is uniformly set to 50,000 with a batch size of 2 for every case. All experiments were conducted on 4 NVIDIA(R) V100 GPUs.

\subsection{Results}

\renewcommand\arraystretch{1}
\begin{table*}
\caption{
User study results over CelebA-HQ \cite{karras2017progressive}, AFHQ \cite{choi2020stargan} Dog and Cat. The \textit{Edit Accuracy} refers to whether the manipulations faithfully reflect the target text descriptions, and \textit{Visual Realism} denotes the photorealism of the manipulated image. The numbers in the table are average rankings of users' preferences, the lower the better.
}
\label{tab_us}
\begin{center}
\scalebox{1.0}{
\begin{tabular}{|c|cc|cc|cc|}
\hline
\multirow{2}{*}{\textbf{Methods}} 
& \multicolumn{2}{c|}{\textbf{CelebA-HQ\cite{karras2017progressive}}} 
& \multicolumn{2}{c|}{\textbf{AFHQ Dog \cite{choi2020stargan}}}
& \multicolumn{2}{c|}{\textbf{AFHQ Cat \cite{choi2020stargan}}}\\
\cline{2-7}
& Edit Accuracy $\downarrow$ & Visual Realism $\downarrow$ & Edit Accuracy $\downarrow$ & Visual Realism $\downarrow$ & Edit Accuracy $\downarrow$ & Visual Realism $\downarrow$ \\
\hline
TediGAN \cite{xia2021tedigan} & 3.23 & 2.99 & N/A & N/A & N/A& N/A\\
StyleCLIP \cite{patashnik2021styleclip} & 2.73 & 2.51 & 2.40 & 2.29 & 2.46 & 2.28 \\
StyleGAN-NADA \cite{gal2021stylegan} & 2.35 & 2.60 & 1.94 & 1.99 & 1.82 & 1.97 \\
\hline
Ours & \textbf{1.69} & \textbf{1.89} & \textbf{1.66} & \textbf{1.71} & \textbf{1.71} & \textbf{1.74} \\
\hline
\end{tabular}}
\end{center}

\end{table*}  

\begin{figure}[t]
    \centering
    \includegraphics[width=1.0\linewidth]{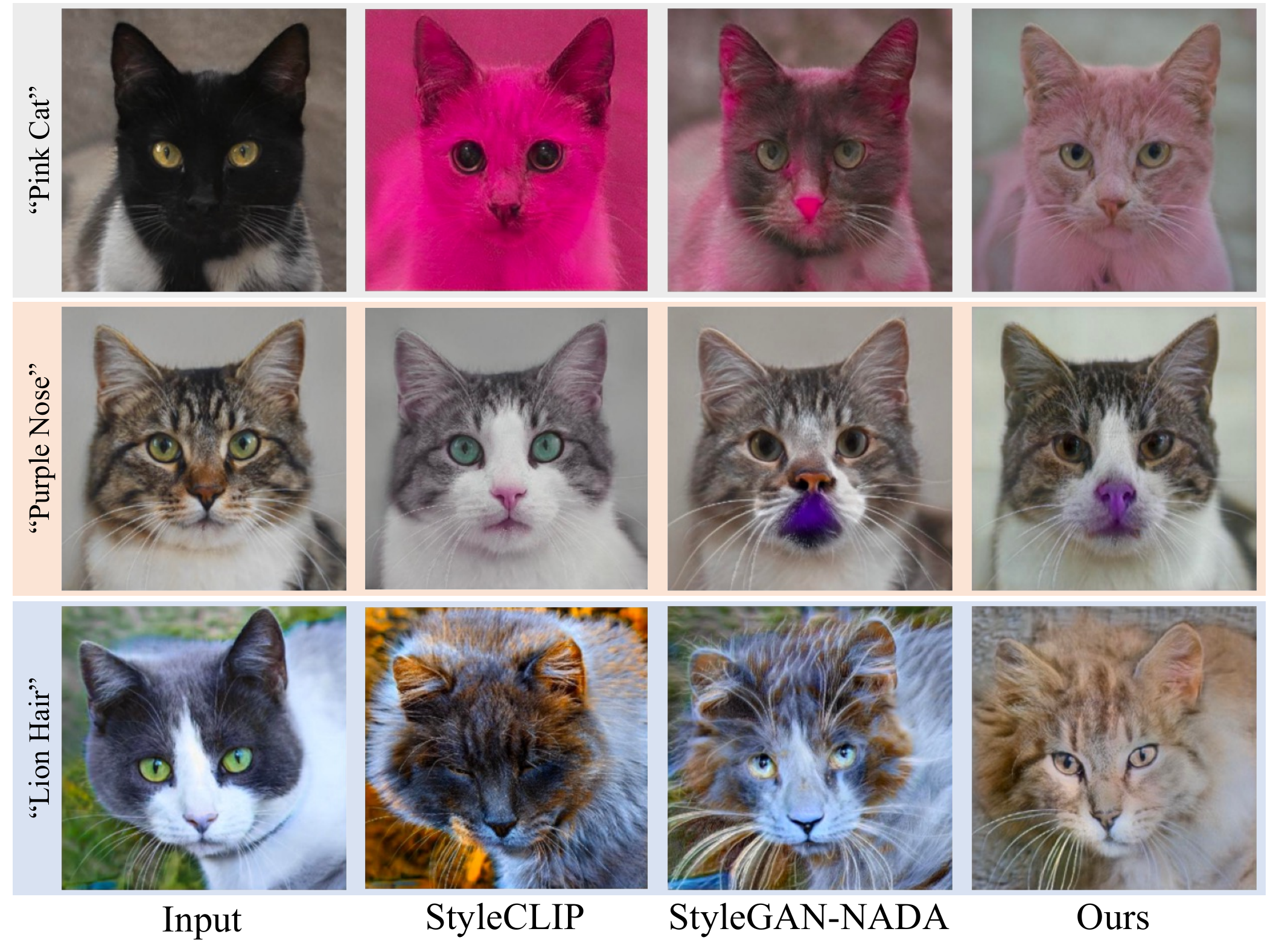}
    \caption{Visual comparison with the state-of-the-art over AFHQ Cat \cite{choi2020stargan}. The target texts appear on the left column and the input images are sampled randomly.}
    \label{comp_cat}
\end{figure}

Figs. \ref{comp_face}, \ref{comp_dog} and \ref{comp_cat} show qualitative results over CelebA-HQ \cite{karras2017progressive}, AFHQ \cite{choi2020stargan} Dog and Cat, respectively. More illustrations and quantitative analysis are shared in Supplementary Materials.

We first compare CF-CLIP with TediGAN \cite{xia2021tedigan}, StyleCLIP \cite{patashnik2021styleclip} and StyleGAN-NADA \cite{gal2021stylegan} on CelebA-HQ \cite{karras2017progressive} with counterfactual manipulation on facial images. The driving texts are designed to cover both global (e.g., "Cyberpunk") and local (e.g., "Elf Ears") manipulations. As shown in Fig. \ref{comp_face}, TediGAN \cite{xia2021tedigan} tends to produce artifacts instead of meaningful changes toward the target. StyleCLIP \cite{patashnik2021styleclip} struggles to yield promising counterfactual manipulation especially for local manipulations, e.g. it edits the background and cloth instead of the target regions for "Purple Eyebrows" and "Green Lipstick". 
StyleGAN-NADA enables out-of-domain manipulation and its manipulations generally meet the target descriptions, but it often suffers from degradation in both identity and image quality even after fine-tuning with StyleCLIP mapper.
CF-CLIP instead generates faithful and high-fidelity manipulations with minimal identity loss for various counterfactual text descriptions.

We also compare CF-CLIP with StyleCLIP \cite{patashnik2021styleclip} and StyleGAN-NADA \cite{gal2021stylegan} on AFHQ \cite{choi2020stargan} Dog and Cat domains. As shown in Fig. \ref{comp_dog}, StyleCLIP \cite{patashnik2021styleclip} hardly meets any counterfactual descriptions and tends to produce unwanted semantic changes.
StyleCLIP-NADA demonstrates promising editing directions toward target texts, but the image quality and object semantics are degraded. In contrast, CF-CLIP yields faithful and visually appealing manipulation aligning with the target texts, with great preservation of image quality and identity information. For the AFHQ Cat domain, Fig. \ref{comp_cat} suggests that StyleCLIP \cite{patashnik2021styleclip} makes global changes disregarding the target editing regions. StyleGAN-NADA can produce reasonable manipulation towards the target descriptions, but suffers from excessive or inaccurate editing. Despite slight identity loss, CF-CLIP editing faithfully aligns with the target descriptions with fair realism.

\subsection{User Study}
\begin{figure*}[t]
    \centering
    \includegraphics[width=0.96\linewidth]{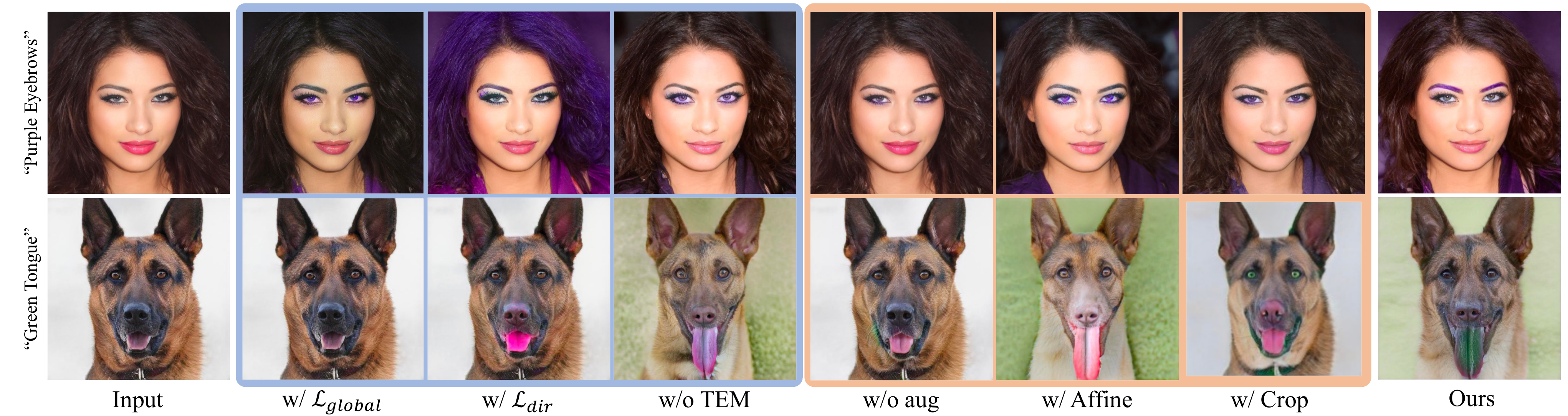}
    \caption{
    Qualitative ablation studies of CF-CLIP over CelebA-HQ \cite{karras2017progressive} and AFHQ Dog \cite{choi2020stargan}: The part in the blue box shows the effectiveness of the proposed CLIP-NCE loss and TEM. The part in the orange box shows the effectiveness of our augmentation scheme. The last column shows the CF-CLIP manipulation. }
    \label{fig_ablation}
\end{figure*}

We also performed subjective user studies to compare the proposed CF-CLIP with state-of-the-art text-guided manipulation methods. The studies were performed over datasets CelebA-HQ \cite{karras2017progressive}, AFHQ \cite{choi2020stargan} Dog \& Cat. 
For each dataset, we randomly sampled 3 manipulated images from each of 8 different counterfactual editing descriptions for each benchmarking method, and then shuffled their orders, which yields 24 groups of results for evaluation.
The images are then presented to 30 recruited volunteers who are separately asked for two tasks. The first task is to measure \textit{Edit Accuracy}, where the subjects need to provide the ranking for each method based on how well the manipulated images by different methods meet the target descriptions. The second task requires the subjects to rank the same images according to their \textit{Visual Realism}, i.e., how naturally the counterfactual concepts fit with the manipulated images. For both tasks, 1 means the best and 4 the worst (3 for AFHQ Cat and Dog as TediGAN \cite{xia2021tedigan} is only for face-related manipulation). Table \ref{tab_us} shows experimental results. It can be observed that CF-CLIP outperforms the state-of-the-art consistently in both tasks over different datasets.

\subsection{Ablation Study}
We study the individual contributions of our technical designs through several ablation studies as illustrated in Fig. \ref{fig_ablation}. More ablation studies and analysis can be found in the Supplementary Materials.

\subsubsection{CLIP-NCE and TEM} We trained three models to examine the proposed CLIP-NCE loss and TEM module: 1) \textit{w/ $\mathcal{L}_{global}$} that replaces CLIP-NCE loss with the global CLIP loss \cite{patashnik2021styleclip}; 2) \textit{w/ $\mathcal{L}_{dir}$} that replaces CLIP-NCE loss with the directional CLIP loss \cite{gal2021stylegan}; 3) \textit{w/o TEM} that removes the proposed TEM module. As shown in Fig. \ref{fig_ablation}, using global CLIP loss $\mathcal{L}_{global}$ tends to manipulate the color of eyes instead of the target eyebrows in the "Purple Eyebrows" case, and gets stuck in local minimum in the second case. The directional CLIP loss $\mathcal{L}_{dir}$ leads to excessive editing of all related areas in the first case, and fails to edit the tongue color in the second case. Without TEM, the model fails to locate target areas accurately, e.g., it edits the eye color instead of the eyebrows. The ablation studies show that both CLIP-NCE loss and TEM module are indispensable for achieving accurate and realistic counterfactual manipulation.

\subsubsection{Augmentation} We also trained three models to examine the effect of perspective augmentation: 1) \textit{w/o aug} that removes the augmentation scheme
and calculates the loss directly on the generated images; 2) \textit{w/ Affine} the replaces the random perspective augmentation with random affine augmentation; 3) \textit{w/ Crop} that performs random crop augmentation and then resizes images back to their original resolution.
The augmentation details are provided in the Supplementary Materials. 
Without perspective augmentation, the model loses certain understanding of the 3D structure of target object/face and fails to locate the target regions (e.g., eyeshadow in first case and dog mouth in second one). 2D affine transformations on the manipulated images cannot fully substitute the effects of perspective transformations as illustrated in both cases. We can also observe that the random cropping has very limited effects, which tends to lead to undesired editing (e.g., green eyes for the second case) due to the increased local views from random cropping.

\section{Conclusion}
This paper presents CF-CLIP, a novel text-guided image manipulation framework that achieves realistic and faithful counterfactual editing by leveraging the great potential of CLIP. To comprehensively explore the rich semantic information of CLIP for counterfactual concepts, we propose a novel contrastive loss, CLIP-NCE to guide the editing from different perspectives based on predefined CLIP-space directions. In addition, a simple yet effective text embedding mapping (TEM) module is designed to explicitly leverage the CLIP embeddings of target text for more precise editing on latent codes in relevant generator layers. Extensive experiments show that our approach could produce realistic and faithful editing given the target text with counterfactual concepts. Moving forward, more challenging counterfactual manipulations with drastic semantic changes will be explored to further unleash the user's creativity.
\newpage
\bibliographystyle{ACM-Reference-Format}
\bibliography{reference}

\end{document}